\documentclass[sigconf]{acmart}
\usepackage{multirow}
\usepackage{colortbl}
\usepackage{pifont}

\AtBeginDocument{%
  \providecommand\BibTeX{{%
    \normalfont B\kern-0.5em{\scshape i\kern-0.25em b}\kern-0.8em\TeX}}}

\renewcommand\footnotetextcopyrightpermission[1]{} % removes footnote with conference information in first column
\setcopyright{none}

\begin{document}

\title{Enhancing Multi-task Learning Capability of Medical Generalist Foundation Model via Image-centric Multi-annotation Data}

\author{Xun Zhu}
% ~\textsuperscript{\dag}
\authornote{Equal contribution.}
\affiliation{%
  \institution{Department of Electronic Engineering\\ Tsinghua University}
  \city{Beijing}
  \country{China}
  }
\email{zhu-x24@mails.tsinghua.edu.cn}

\author{Fanbin Mo}
\authornotemark[1]
\affiliation{%
  \institution{School of Artificial Intelligence \\ BUPT}
  \city{Beijing}
  \country{China}
  }
\email{mofanbin@bupt.edu.cn}

\author{Zheng Zhang}
\affiliation{%
  \institution{Department of Electronic Engineering \\ Tsinghua University}
  \city{Beijing}
  \country{China}
  }
\email{zzhang24@mails.tsinghua.edu.cn}

\author{Jiaxi Wang}
\affiliation{%
  \institution{Department of Electronic Engineering \\ Tsinghua University}
  \city{Beijing}
  \country{China}
  }
\email{wjx20@mails.tsinghua.edu.cn}

\author{Yiming Shi}
\affiliation{%
  \institution{Department of Electronic Engineering \\ Tsinghua University}
  \city{Beijing}
  \country{China}
  }
\email{sym23@mails.tsinghua.edu.cn}

\author{Ming Wu}
\affiliation{%
  \institution{School of Artificial Intelligence \\ BUPT}
  \city{Beijing}
  \country{China}
  }
\email{wuming@bupt.edu.cn}

\author{Chuang Zhang}
\affiliation{%
  \institution{School of Artificial Intelligence, BUPT \\
  Beijing Wuzi University}
  \city{Beijing}
  \country{China}
  }
\email{zhangchuang@bupt.edu.cn}

\author{Miao Li}
\authornote{Corresponding authors.} % 含义是使用第一个\authornote的标记
\affiliation{%
  \institution{Department of Electronic Engineering \\ Tsinghua University}
  \city{Beijing}
  \country{China}
  }
\email{miao-li@tsinghua.edu.cn}

\author{Ji Wu}
% ~\textsuperscript{\ding{41}}
\authornotemark[2]
\affiliation{%
  \institution{Department of Electronic Engineering \\ College of AI, Tsinghua University}
  \city{Beijing}
  \country{China}
  }
\email{wuji_ee@tsinghua.edu.cn}

\begin{abstract}
The emergence of medical generalist foundation models has revolutionized conventional task-specific model development paradigms, aiming to better handle multiple tasks through joint training on large-scale medical datasets.
However, recent advances prioritize simple data scaling or architectural component enhancement, while neglecting to re-examine multi-task learning from a data-centric perspective.
Critically, simply aggregating existing data resources leads to decentralized image-task alignment, which fails to cultivate comprehensive image understanding or align with clinical needs for multi-dimensional image interpretation.
In this paper, we introduce the image-centric multi-annotation X-ray dataset (IMAX), the first attempt to enhance the multi-task learning capabilities of medical multi-modal large language models (MLLMs) from the data construction level.
To be specific, IMAX is featured from the following attributes:
1) \textit{High-quality data curation.}
A comprehensive collection of more than 354K entries applicable to seven different medical tasks.
2) \textit{Image-centric dense annotation.}
Each X-ray image is associated with an average of 4.10 tasks and 7.46 training entries, ensuring multi-task representation richness per image.
Compared to the general decentralized multi-annotation X-ray dataset (DMAX), IMAX consistently demonstrates significant multi-task average performance gains ranging from 3.20\% to 21.05\% across seven open-source state-of-the-art medical MLLMs.
Moreover, we investigate differences in statistical patterns exhibited by IMAX and DMAX training processes, exploring potential correlations between optimization dynamics and multi-task performance.
Finally, leveraging the core concept of IMAX data construction, we propose an optimized DMAX-based training strategy to alleviate the dilemma of obtaining high-quality IMAX data in practical scenarios.
Related resources will be released soon.

\end{abstract}


\maketitle

\begin{figure*}[t]
\centerline{\includegraphics[scale=0.35]{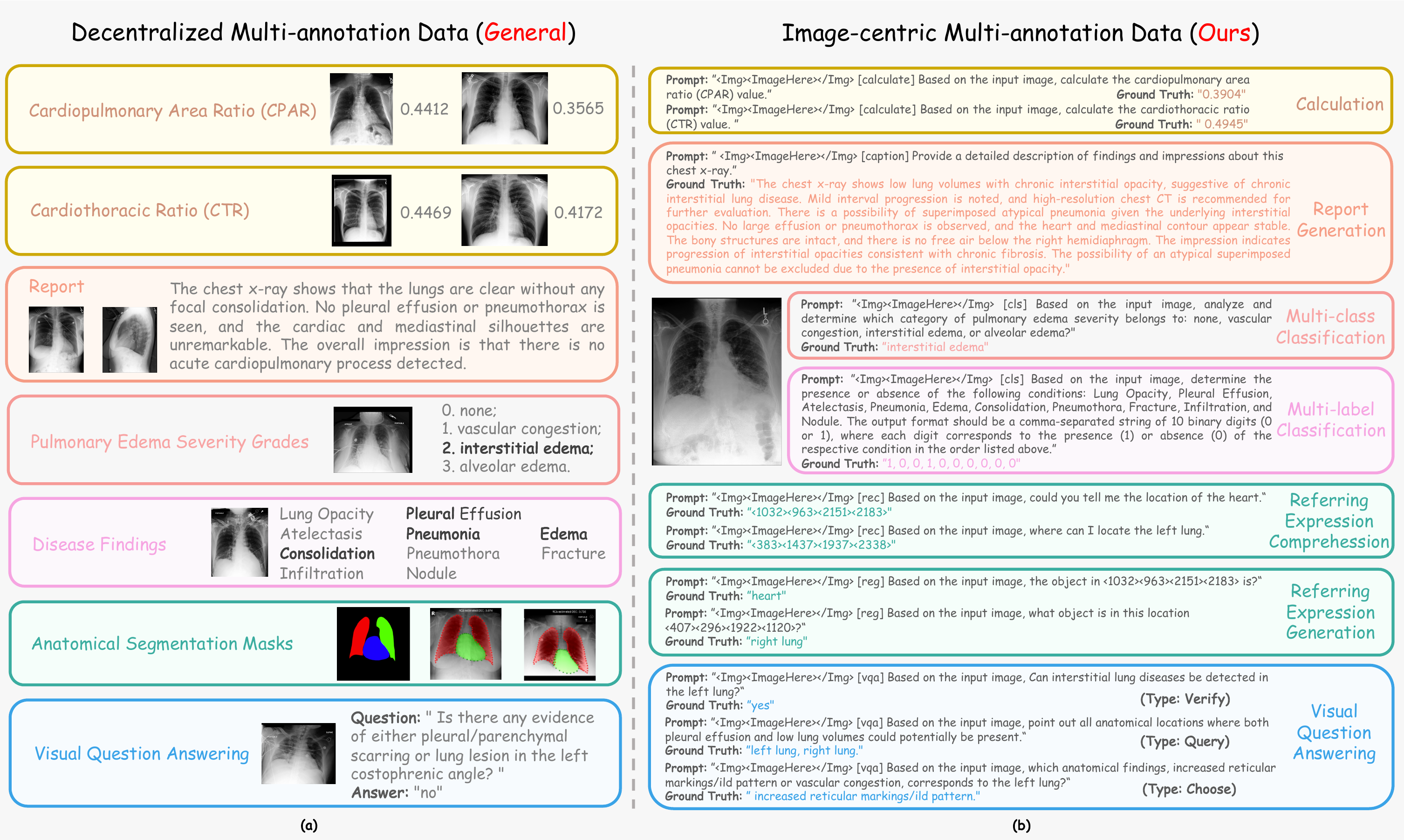}}
\caption{
Illustration of the dataset construction.
(a) Example of general decentralized multi-annotation data for multi-task learning.
(b) Example of our image-centric multi-annotation data for multi-task learning.}
\label{f1}
\end{figure*}

\section{Introduction}
The rapid advancement of multi-modal Large Language Models (MLLMs) has catalyzed transformative progress across diverse domains.
By leveraging large-scale multimodal data, MLLMs combine visual perception with language understanding, demonstrating the ability to capture intricate visual-textual patterns~\cite{2024A, li2024continuous, liang2024survey}.
In the medical domain, this evolution has primarily spurred the development of specialized MLLMs tailored for task-specific applications, such as medical visual question answering (VQA)~\cite{li2023llava} and radiology report generation~\cite{thawakar2024xraygpt}.
Advancing towards generalist medical AI~\cite{moor2023foundation}, medical generalist foundation models have gone a step further, employing joint training with vast and diverse datasets to holistically handle multiple tasks and modalities within a unified architecture with shared parameters~\cite{zhu2024uni, wu2023towards, tu2024towards}.

To fully unleash the multi-task learning capabilities of medical generalist foundation models, recent efforts have predominantly focused on extensive data scaling in the pretraining and fine-tuning stages~\cite{xie2024medtrinity, chen2024chexagent, li2024gmai}, coupled with targeted architectural refinements~\cite{zhu2024uni, huang2024towards}. 
However, a critical perspective remains underexplored: what type of data is conducive to model optimization in the scenario of multi-task learning?

In this study, we investigate the above question by establishing multi-task learning scenario involving chest X-rays, which are the most frequently performed imaging tests in clinical practice and medical imaging research~\cite{chen2024chexagent, johnson2019mimic}.
The inherent richness of visual information and fine-grained semantics in X-rays makes them a powerful tool from disease diagnosis to treatment planning~\cite{rajpurkar2017chexnet, topol2019high}.
Despite the potential of X-rays to support various medical tasks, the construction methods of existing multi-task learning datasets fail to take advantage of this potential.

On the one hand, while combining existing X-ray datasets to complete different tasks seems to be a simple and convenient way to construct multi-granularity annotation data, they suffer from a critical limitation: the image collections for different tasks are decentralized.
This decentralization leads to insufficient utilization of image data, ignoring the collaborative promotion of a single medical image completing multiple tasks, ultimately restricting the ability of models to develop a holistic understanding of each medical image.
On the other hand, in real-world clinical scenarios, a single X-ray image is typically subjected to multiple queries and interpretations, each addressing different aspects of the patient's condition.
However, the current training and testing paradigms treat these tasks as independent, introducing potential biases that hinder cross-task generalization and multi-dimensional image interpretation.

To address these data-centric limitations, we innovatively propose the concept of image-centric multi-annotation data and introduce the \textbf{I}mage-centric \textbf{M}ulti-\textbf{A}nnotation \textbf{X}-ray (IMAX) dataset.
Different from the general decentralized multi-annotation X-ray dataset (DMAX), IMAX is structured around a centralized image-centric framework, where each X-ray image is annotated with multiple labels and textual descriptions that span a wide range of medical tasks.
Specifically, IMAX is characterized by two pivotal attributes:
1) High-quality data curation.
All raw data come from recognized authoritative websites (e.g., PhysioNet~\cite{goldberger2000physiobank}) and widely used open-source datasets.
After strict and meticulous data cleaning, a comprehensive collection of 47,600 unique X-ray images and 354,595 entries can be applied to seven medical tasks including calculation, report generation, multi-class classification, multi-label classification, referring expression comprehension (REC), referring expression generation (REG), and VQA.
2) Image-centric annotation density.
Each imaging study achieves an average of 4.10 tasks and 7.46 training entries per X-ray image, surpassing DMAX by factors of 3.3× and 3.6× respectively.
This construction method ensures that the dataset captures rich and interconnected semantics of medical images, enabling foundation models to learn from the unified and comprehensive representation of the data.
To our knowledge, IMAX represents the first highly descriptive multi-modal collection specifically engineered to enhance the multi-task learning capabilities of medical foundation models.

To substantiate our argument, we conduct straightforward and comprehensive experiments.
We select and evaluate seven open-source state-of-the-art medical MLLMs, including LLaVA-Med~\cite{li2023llava}, Xray-GPT~\cite{thawakar2024xraygpt}, MiniGPT-Med~\cite{alkhaldi2024minigpt}, Uni-Med~\cite{zhu2024uni}, LLaVA-Tri~\cite{xie2024medtrinity}, MedM-VL~\cite{shi2025medm}, and HealthGPT~\cite{lin2025healthgpt}, comparing their multi-task performance after fine-tuning with IMAX and DMAX, respectively.
Despite differences in model architectures, training parameter configurations, and medical pretraining data sources, the comparative results indicate that IMAX consistently yields significant enhancements in multi-task abilities, with average performance gains spanning 3.20\%-21.05\%.
In addition, we observe and analyze the differences in model training using IMAX and DMAX data under a unified and standard MLLM architecture.
Specifically, we quantify the spectral entropy of non-zero eigenvalues and dominant eigenvalue ratio of the Fisher information matrix (FIM), which are closely related to the stability and convergence speed of the optimization process~\cite{martens2020new,advani2020high}.
Through the statistical patterns presented at various stages of the training process, we are committed to exploring the potential correlations between optimization dynamics and multi-task performance.
Moreover, to address the practical challenge of high acquisition costs of IMAX data, we propose an optimized three-stage DMAX-based training strategy.
Drawing inspiration from semi-supervised learning and the core concept of IMAX data construction, our strategy achieves a balance between annotation quality and model performance by generating pseudo IMAX data.
We hope our study can significantly motivate more researchers to focus on enhancing multi-task learning capability of medical generalist foundation model from data perspective, while promoting the core concept of image-centric multi-annotation data to broader domains.
In a nutshell, the contributions of this paper can be summarized as
follows:

\textbullet\; We present IMAX, the first image-centric multi-annotation X-ray dataset, which can perform seven medical tasks including calculation, report generation, multi-class classification, multi-label classification, referring expression comprehension, referring expression generation, and visual question answering.

\textbullet\; We conduct extensive comparisons across seven open-source state-of-the-art medical MLLMs, demonstrating that our IMAX achieves significant enhancements in multi-task learning capabilities compared to the general DMAX.

\textbullet \; Focusing on how IMAX optimizes multi-task learning during the training process, we provide statistical patterns and instructive findings from the perspective of Fisher information matrix.

\textbullet \; Building on the core principles of IMAX data construction, we develop an optimized DMAX training strategy to mitigate the practical challenge of acquiring high-quality IMAX data.

\section{Related Work}
\subsection{Medical Multi-modal Large Language Model}
The rapid evolution of LLMs has catalyzed advancements in MLLMs, which have garnered substantial research attention due to their powerful representational capacity and remarkable proficiency in multi-modal data processing~\cite{liang2024survey}.
The development of medical MLLMs aims to create an efficient medical assistant, following the architecture design and training strategy of MLLMs in the general field, using a large amount of medical data for pretraining and instruction fine-tuning~\cite{lin2024has}.
For example, LLaVA-Med~\cite{li2023llava}, XrayGPT~\cite{thawakar2024xraygpt}, MedBLIP~\cite{chen2024medblip}, and Med-Flamingo~\cite{moor2023med} are the adaptions of MiniGPT-4, LLaVA, BLIP-2 and Flamingo within the medical field, respectively. 
Usually, medical MLLMs involve a stage of individual fine-tuning based on specific tasks or specific modalities data.
RaDialog~\cite{pellegrini2023radialog}, MAIRA-1~\cite{hyland2023maira}, and LLaVA-Rad~\cite{chaves2024training} are optimized for 2D chest X-ray analysis, while Dia-LLaMA~\cite{chen2024dia} and BrainGPT~\cite{li2025towards}  are pioneers designed for 3D CT report generation.
PathChat~\cite{lu2023foundational} and PathAsst~\cite{sun2024pathasst} are developed for human pathology.
SkinGPT-4~\cite{zhou2024pre} uses the collection of skin disease images along with clinical concepts and doctors’ notes for dermatological diagnosis.
Med-MLLM~\cite{liu2023medical} has fine-tuned three downstream COVID-19 decision support tasks, including reporting, diagnosis, and prognosis.
BiomedGPT~\cite{zhang2024generalist} utilizes pretrained knowledge within a unified framework to specialize effectively on 25 datasets through fine-tuning, respectively.
While remaining grounded in established task-specific paradigms, medical MLLMs derive their performance superiority over traditional methods by expanding the training corpus and scaling the parameter architectures.

\subsection{Medical Generalist Foundation Model}
With the proposal of the concept of generalist medical artificial intelligence (GMAI)~\cite{moor2023foundation}, medical generalist foundation model aims to perform various tasks within a unified parameter sharing architecture.
It seeks to eliminate the reliance on task-specific modules and further fine-tuning, thereby disrupting the traditional task-specific approach to model development.
Toward the ambitious goals mentioned above,
LLM-CXR~\cite{lee2023llm} achieves bidirectional understanding and generation of chest X-rays through joint training across four primary tasks distinguished by instructions.
Med-PaLM M~\cite{tu2024towards} is a closed-source generalist biomedical AI system, which can handle 14 diverse tasks in multiple modalities.
MiniGPT-Med~\cite{alkhaldi2024minigpt} is capable of handling 3 tasks from radiological images including X-rays, CT scans, and MRIs.
RadFM~\cite{wu2023towards} allows the integration of text input with 2D or 3D medical scans, and generates responses for diverse radiologic tasks.
MedM-VL~\cite{shi2025medm} constructs two different models for 2D and 3D modalities, both of which support general-purpose medical tasks and domain-specific fine-tuning.
M3D-LaMed~\cite{bai2024m3d} is designed for various 3D medical tasks.
Uni-Med~\cite{zhu2024uni} provides a superior solution to the tug-of-war problem of multi-task learning by introducing mixture-of-experts (MoE) into the connector.
MedPLIB~\cite{huang2024towards} introduces MoE into the LLM, which effectively coordinates multi-task learning.
HealthGPT~\cite{lin2025healthgpt} presents H-LoRA, an optimized multi-LoRA PEFT architecture based on task-gated decoupling, to effectively mitigate data conflict issues.
However, recent advances prioritize simple data scaling or architectural component enhancement, while overlooking to re-examine multi-task learning from a deeper data-centric perspective.

\subsection{Multi-granularity Annotation Dataset}
% To enhance the multi-task learning efficiency and ability of the medical generalist foundation model, 
To empower the medical generalist foundation model with comprehensive multi-modal multi-task understanding ability, the training data prioritize wider coverage, higher quality, and richer granularity.
MedTrinity-25M~\cite{xie2024medtrinity} establishes a pretraining dataset across 10 modalities, annotated with text ranging from global attributes (modality and organ detection) to local information (ROI analysis, lesion texture, and region-wise correlations).
GEMeX~\cite{liu2024gemex} is a groundable and explainable medical VQA benchmark for chest X-ray diagnosis, which supports 4 clinically-relevant question types through visual-textual justification mechanisms.
GMAI-VL-5.5M~\cite{li2024gmai} covers 13 medical imaging modalities and 18 specialties by converting hundreds of specialized medical datasets into meticulously constructed image-text pairs.
CheXinstruct~\cite{chen2024chexagent} is an instruction-tuning dataset with 6M instruction-image-answer triplets that aggregate from 34 tasks and 65 unique datasets, spanning categories including
coarse and fine-grained image understanding, question answering, and text generation.
For 3D chest CT interpretation, RadGenome-Chest CT~\cite{zhang2024radgenome} introduces region-guided reasoning through anatomically grounded reports and visual-evidence VQA pairs, anchored to organ-level segmentation masks.
In comparison, our IMAX is a pioneering work that explores enhancing multi-task learning capabilities of medical generalist foundation model by constructing image-centric multi-annotation data, which can perform seven medical tasks including calculation, report generation, multi-class classification, multi-label classification, REC, REG, and VQA.

\section{Image-centric Multi-annotation X-ray Dataset}
In this section, we introduce the image-centric multi-annotation X-ray dataset, the first image-text collection to cover multi-task descriptions around each X-ray, which aims to enhance the model's ability for multi-task joint training.
The whole process of data collection and processing is elaborated in Section~\ref{3.1}, while dataset statistics and comparison are presented in Section~\ref{3.2}.
\subsection{Data Collection and Processing}
\label{3.1}
As a large publicly available JPG format chest radiographs dataset, MIMIC-CXR-JPG~\cite{johnson2019mimic} is widely used in medical research.
In addition to its accompanying structured radiology reports, an increasing number of studies have utilized various expert models to derive diverse types of data from images within MIMIC-CXR-JPG, tailored to meet specific research needs.

In this study, we have first collected a substantial amount of derived data related to X-ray images, as illustrated in Figure~\ref{f1}(a).
Specifically, CXR-Cardiomegaly~\cite{duvieusart2022multimodal} provides two types of cardiome-galy biomarker values extracted from images: cardiothoracic ratio (CTR) and cardiopulmonary area ratio (CPAR).
MIMIC-CXR-PE-Severity~\cite{horng2021deep} offers four distinct severity grades of pulmonary edema.
Each image in CXR-LT~\cite{holste2024towards} is labeled with one or multiple labels from a set of 45 disease findings by parsing radiology reports.
CheXmask~\cite{gaggion2024chexmask} provides anatomical segmentation masks for the heart, left lung, and right lung, which can be further processed into bounding boxes.
MIMIC-Ext-MIMIC-CXR-VQA~\cite{bae2023ehrxqa} features questions generated using templates developed under the guidance of medical experts, addressing both standard content from previous medical VQA and more complex scenarios involving set and logical reasoning.
For more details on data sources and availability, see Supplementary Materials A.

We refer to the collected raw image-text pairs as decentralized multi-annotation data.
While directly utilizing the above data can fulfill the requirements of multi-task learning for X-rays—a generally adopted approach—it is noteworthy that the intersection of image sets corresponding to different tasks is relatively small.
This limited overlap may pose challenges for achieving comprehensive multi-task learning across diverse annotations.

As illustrated in Figure~\ref{f1}(b), we are committed to preparing image-centric multi-annotation data.
For this purpose, we have designed different prompt templates for seven tasks and accompanied by examples, including calculation, report generation, multi-class classification, multi-label classification, referring expression comprehension, referring expression generation, and visual question answering.
Our objective is to ensure that, for each X-ray, the data cover as many different tasks as possible, thereby maximizing the utility and versatility of the image in the scenario of multi-task learning. 

\begin{figure}[!]
\centerline{\includegraphics[scale=0.44]{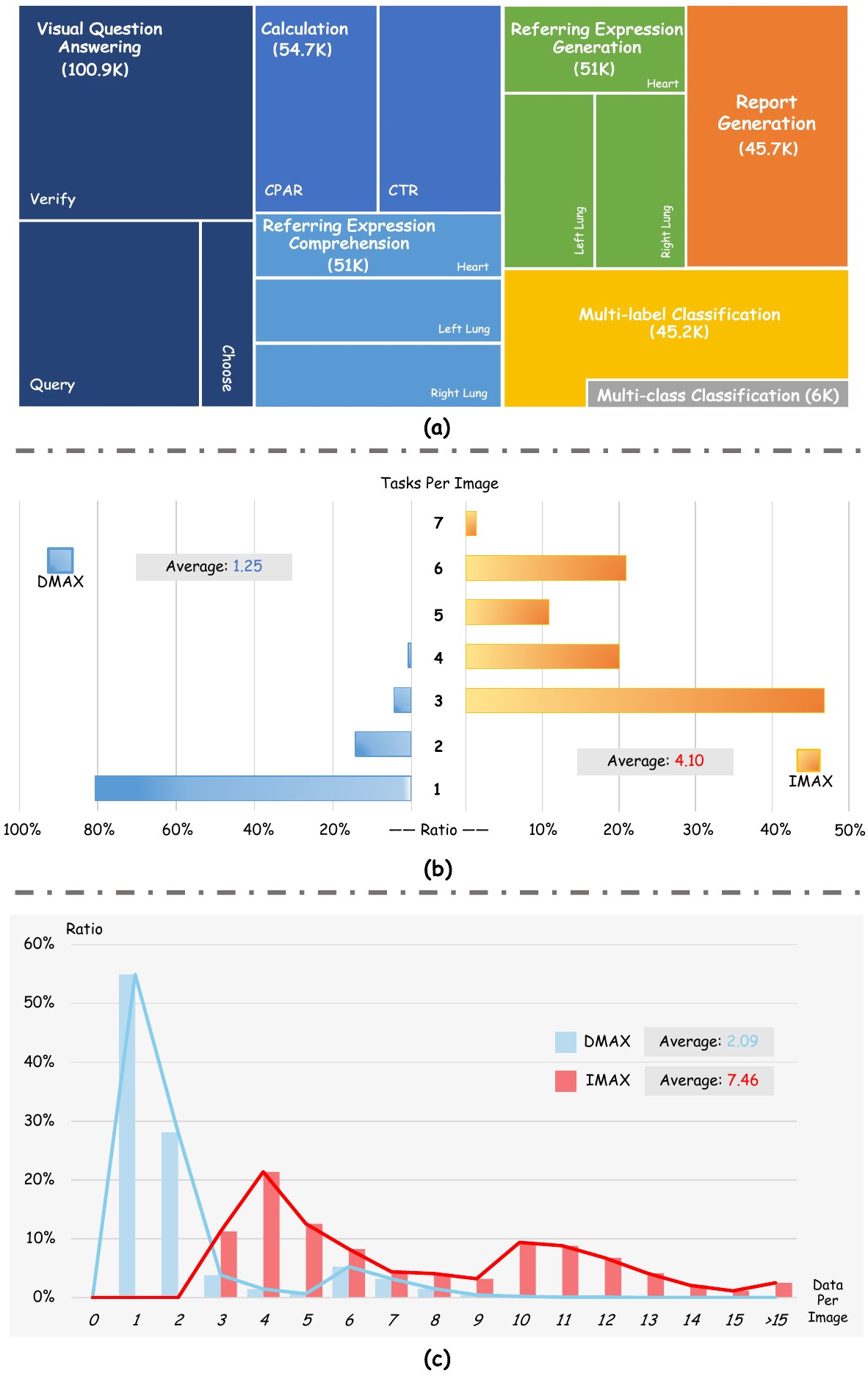}}
\caption{
Dataset statistics.
(a) The overall composition of image-centric multi-annotation X-ray dataset (IMAX).
(b) The distribution of the number of tasks corresponding to each image.
(c) The distribution of the number of train data entries corresponding to each image.}
\label{f2}
\end{figure}

\begin{figure*}[t]
\centerline{\includegraphics[scale=0.361]{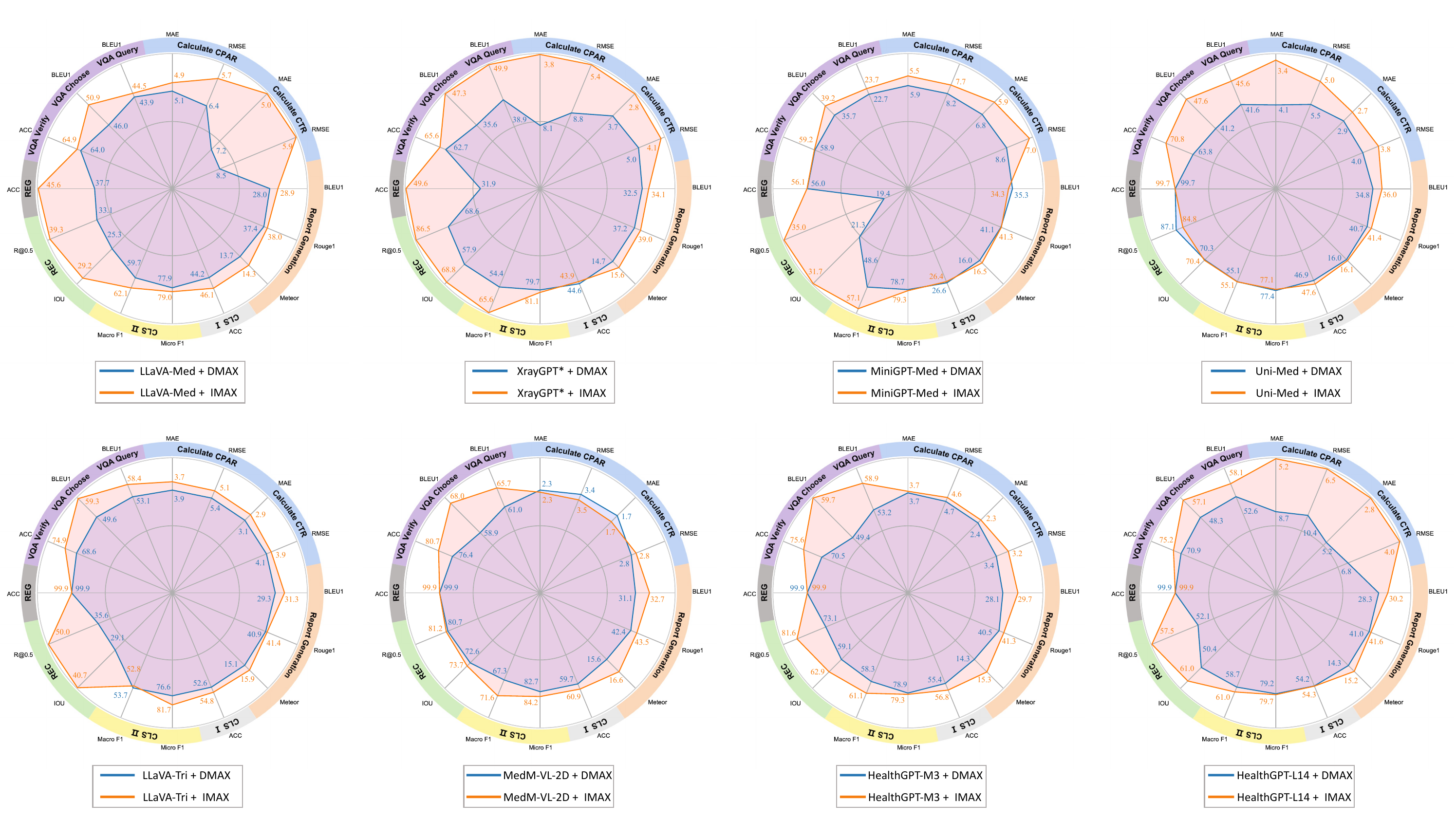}}
\caption{Illustration of the benefits image-centric multi-annotation data bring to the multi-task learning ability of medical foundation model.
We compare the performance of various medical MLLMs fine-tuned with IMAX and DMAX, respectively.
CLS \uppercase\expandafter{\romannumeral1} and CLS \uppercase\expandafter{\romannumeral2}  represent multi-class classification and multi-label classification, respectively.}
\label{f3}
\end{figure*}

\subsection{Dataset Statistics}
\label{3.2}
Through meticulous data filtering and cleaning, we ultimately present our image-centric multi-annotation X-ray dataset (IMAX) and describe the overall composition of IMAX in Figure~\ref{f2}(a).
Specifically, IMAX comprises a total of 47,600 unique X-rays and 354,595 data entries, distributed as follows: 100,901 for VQA, 54,684 for calculation, 51,045 for REC, 51,045 for REG, 45,715 for report generation, 45,186 for multi-label classification, and 6,019 for multi-class classification.
We partition IMAX into train and test sets with a ratio of 4:1, resulting in 38,077 images and 284,017 data entries allocated for training.

To facilitate intuitive comparison with general multi-task learning datasets, we rigorously sample data from the remaining decentralized multi-annotation data to match the task-specific data volumes in the IMAX train set.
This process results in the creation of the decentralized multi-annotation X-ray dataset (DMAX), which comprises 136,218 unique X-rays.
In Figure~\ref{f2}(b), we present the distribution of the number of tasks corresponding to each image.
Over 80\% of the images in DMAX are associated with only a single task, with an average of 1.25 tasks per image.
In contrast, images in IMAX can be used for 3 to 7 tasks, with an average of 4.10 tasks per image.
In Figure~\ref{f2}(c), we present the distribution of the number of train data entries corresponding to each image.
DMAX and IMAX exhibit distinct distribution patterns: Each image in DMAX corresponds to an average of 2.09 train data entries, whereas the number increases significantly to 7.46 in IMAX.
The statistical results highlight the differences in data density and annotation coverage between the two datasets constructed under distinct methodologies.

\section{Experiments}
In this section, we demonstrate the effectiveness of utilizing IMAX for multi-task learning through experiments and analyses.
We perform extensive comparisons of several open-source state-of-the-art medical MLLMs fine-tuned with IMAX and DMAX in Section~\ref{4.1}.
Moreover, in Section~\ref{4.2}, we present observations and analysis within a unified and standard MLLM framework, with the aim of providing insight into the possible underlying mechanisms behind performance improvements and developing an optimized DMAX training strategy to mitigate the practical dilemma of acquiring high-quality IMAX data.
\subsection{Extensive Comparison}
\label{4.1}

\subsubsection{Implementation Details}
\paragraph{Baseline Settings}
We select seven open-source state-of-the-art medical MLLMs for extensive comparative analysis.
LLaVA-Med~\cite{li2023llava} achieves domain-specific medical VQA through knowledge integration.
XrayGPT~\cite{thawakar2024xraygpt} specializes in chest radiograph summarization.
MiniGPT-Med~\cite{alkhaldi2024minigpt} is designed for heterogeneous radiological vision-language tasks.
Uni-Med~\cite{zhu2024uni} explores multi-modal multi-task joint optimization via mixture-of-experts in the connector.
LLaVA-Tri~\cite{xie2024medtrinity} is fine-tuned on 25 million multi-granularity medical annotations.
MedM-VL-2D~\cite{shi2025medm} demonstrate strong performance across general-purpose medical benchmarks and serve as an effective initialization for fine-tuning.
HealthGPT~\cite{lin2025healthgpt} presents H-LoRA to effectively mitigate data conflict issues and releases two model weights, HealthGPT-M3 and HealthGPT-L14, according to the different selection of LLM.

For all baselines, we initialize with publicly released pre-trained weights (if available) prior to downstream task fine-tuning.
Each model is respectively fine-tuned on DMAX and IMAX through joint multi-task training.
The training protocol utilizes two NVIDIA A800-SXM4-80GB GPUs for 3 epochs, while all other hyperparameters are retained as the original configuration.
For more details on baselines, see
Supplementary Materials B.

\paragraph{Evaluation Metrics}
We report the scaled mean absolute error (MAE) and root mean squared error (RMSE) with a factor of 100 for the calculation task;
BLEU1, Rouge1 and Meteor for the report generation task;
accuracy for the multi-class classification task;
micro F1 and macro F1 for the multi-label classification task;
intersection over union (IOU) and recall at 0.5 IOU threshold (R@0.5) for the REC task;
accuracy for the REG task;
accuracy and BLEU1 for the VQA task.
In addition, we use $\Delta$ to quantify the average performance gains across all tasks, which is defined as follows:
\begin{equation}
\Delta = \frac{1}{S} \textstyle\sum_{i=1}^{S} \left(M_{m,i} - M_{b,i} \right) / M_{b,i} \times 100\% 
\end{equation}
where $S$ denotes the total number of evaluation metrics, $M_{m,i}$ and $M_{b,i}$ are the metrics under IMAX and DMAX, respectively.

\begin{table}[b]
\caption{
The results of the improved metrics ratio and the average gains $\Delta$ across baselines.}
\label{t1}
\renewcommand{\arraystretch}{1.1}
\begin{tabular}{lcc}
\toprule
\multirow{2}{*}{Models} & Improved / Total & \multicolumn{1}{l}{Average Gains} \\
 & Metrics Count & {$\Delta \left(\uparrow\right)$} \\
\midrule
LLaVA-Med~\cite{li2023llava}& 16 / 16 & 10.14\% \\
XrayGPT*~\cite{thawakar2024xraygpt} & 15 / 16 & 21.05\% \\
MiniGPT-Med~\cite{alkhaldi2024minigpt} & 14 / 16 & 12.94\% \\
Uni-Med~\cite{zhu2024uni} & 14 / 16 & 4.81\% \\
LLaVA-Tri~\cite{xie2024medtrinity} & 14 / 16 & 10.17\% \\
MedM-VL-2D~\cite{shi2025medm} & 13 / 16 & 3.20\% \\
HealthGPT-M3~\cite{lin2025healthgpt} & 15 / 16 & 5.47\% \\
HealthGPT-L14~\cite{lin2025healthgpt} & 15 / 16 & 16.92\% \\
\bottomrule
\end{tabular}
\end{table}

\subsubsection{Overall Results} \hfill \\

Figure~\ref{f3} demonstrates the performance comparison of all baselines when trained with DMAX and IMAX through multi-task joint learning.
The radar charts visualize the relative performance improvement ratio achieved by IMAX over DMAX, where a larger ratio corresponds to a wider gap in the radar plot.
To avoid ambiguity, we explicitly annotate the absolute metric values.
Since original XrayGPT~\cite{thawakar2024xraygpt} only makes connector parameters open during training, which is insufficient to support the multi-modal input-output alignment required for multi-task learning, we additionally unfreeze the LLM parameters and  mark with $*$.

Notably, regardless of variations in model architectures, training parameter configurations, or medical pretraining data sources, the utilization of IMAX data consistently demonstrates significant improvements in the multi-task learning performance of medical MLLMs.
In Table~\ref{t1}, we report the count of improved metrics and the average gains $\Delta$ across all metrics for each baseline.
LLaVA-Med~\cite{li2023llava} achieves improvements in all 16 metrics across 7 tasks, with an average performance gains of 10.14\%.
MedM-VL-2D~\cite{shi2025medm} shows improvements in 13 metrics, which is the least among all models, resulting in the lowest average gains of 3.20\%.
Uni-Med~\cite{zhu2024uni} and HealthGPT-M3~\cite{lin2025healthgpt} exhibit relatively low average performance gains (4.81\% and 5.47\%) among all baselines.
We attribute this to its architecture specifically optimized to address optimization conflicts in multi-task joint training, which inherently diminishes the potential gains achievable solely through data level.
In contrast, IMAX achieves substantial performance improvements exceeding 10\% for all other baselines, with the highest gains being 21.05\%.

\subsection{Insightful Observation}
\label{4.2}
\subsubsection{Preliminaries}
\paragraph{Fisher Information Matrix}
As one of the fundamental tools to quantify the sensitivity of model parameters to variations in the data probability distribution, the Fisher information matrix (FIM) plays a pivotal role in fields such as neural network weight pruning and model optimization~\cite{fisher1922mathematical,ly2017tutorial,karakida2019universal,liu2020quantum}.
The FIM of a network $f$ and its set of parameters $\theta \in \mathbb{R}^{p}$ can be defined as follows:
\begin{equation}
FIM(\theta):=\mathbb{E}\left[\nabla_\theta \log p\left(y \mid x ; \theta\right) \cdot \nabla_\theta \log p\left(y \mid x ; \theta\right)^T\right]
\end{equation}
where $x$ and $y$ are the input and output of the network, respectively,
$\nabla_\theta \log p\left(y \mid x ; \theta\right)$ represents the gradient of the log-likelihood with respect to the parameters.
FIM can be estimated by computing the empirical expectation $\mathbb{E}$ of the gradient outer products across the total number of data samples $N$.

\paragraph{Eigenvalue Distribution of FIM}
The eigenvalue distribution of the FIM directly influences the stability and convergence speed of the optimization process~\cite{martens2020new,advani2020high}.
Small eigenvalues indicate lower information content in the corresponding directions, implying that parameter variations have negligible impact on the model output. 
Consequently, parameter updates along these directions contribute relatively little to improving model performance during optimization~\cite{karakida2019universal}. 
Larger eigenvalues signify higher information content in their respective directions, where parameter changes substantially affect the model output~\cite{amari2019fisher}.
Therefore, we examine both the dominant eigenvalue ratio $\rho$ and the spectral entropy $SE$ of non-zero eigenvalues, which are formally defined as:
\begin{equation}
\rho=\frac{\lambda_{\max }}{\sum_{i=1}^d \lambda_i} \times   100\%
\end{equation}
\begin{equation}
SE(\Lambda)=-\sum_{i=1}^d \tilde{\lambda}_i \log \tilde{\lambda}_i, \quad \tilde{\lambda}_i=\frac{\lambda_i}{\sum_j \lambda_j}
\end{equation}
where $\Lambda=\left\{\lambda_i\right\}_{i=1}^d$ is the set of non-zero eigenvalues, $\lambda_{\max}$ is the maximum eigenvalue.

Note that multi-task learning is a profoundly complex and diverse scenario, where systematically assessing the impact of different training data within a fixed architecture remains methodologically challenging.
We seek to identify potential patterns within our experimental scenario by combining the statistical results observed during training with the final evaluation metrics.

\begin{table*}[t]
\small
\tabcolsep= 0.03cm
\renewcommand\arraystretch{1.2}
\caption{Experiment results under our defined unified MLLM architecture.
DMAX \& Pseudo IMAX represents our proposed three-stage strategy for DMAX-based training.
Results in \textbf{bold} denote the overall best, while results with \underline{underlines} indicate the second best.
}
\label{t2}
\begin{tabular}{cccccccccccc}
\toprule
\multicolumn{1}{c}{\multirow{3}{*}{\textbf{Data Selection}}} & \multicolumn{2}{c}{\textbf{Calculation}} & \multirow{2}{*}{\textbf{Report Generation}} & \multicolumn{2}{c}{\textbf{Classification}} & \multirow{2}{*}{\textbf{REC}} & \multirow{2}{*}{\textbf{REG}} & \multicolumn{3}{c}{\textbf{VQA}} & \textbf{Average} \\
\multicolumn{1}{c}{} & \textbf{CPAR} & \textbf{CTR} &  & \textbf{Multi-class} & \textbf{Multi-label} &  &  & \textbf{Verify} & \textbf{Choose} & \textbf{Query} & \textbf{Gains} \\
\cmidrule(r){2-12}
\multicolumn{1}{c}{} & MAE/RMSE & MAE/RMSE & BLEU/ROUGE/Meteor & ACC & Micro/Macro F1 & IOU/ACC & ACC & ACC & BLEU & BLEU & {$\Delta \left(\uparrow\right)$} \\
\midrule
DMAX & \cellcolor[HTML]{F2F2F2}6.93 / 8.22 & \cellcolor[HTML]{F2F2F2}3.67 / 4.96 & \cellcolor[HTML]{F2F2F2}\;33.25\; / \underline{40.34 }  / 15.18 & \cellcolor[HTML]{F2F2F2}26.97 & \cellcolor[HTML]{F2F2F2}73.62 / 47.84 &\cellcolor[HTML]{F2F2F2}50.89 / 63.95 & \cellcolor[HTML]{F2F2F2}64.13 & \cellcolor[HTML]{F2F2F2}62.75 & \cellcolor[HTML]{F2F2F2}40.69 & \cellcolor[HTML]{F2F2F2}\underline{39.06} & \cellcolor[HTML]{F2F2F2}- \\

IMAX & \textbf{3.31} / \textbf{4.76} & \textbf{2.52} / \textbf{3.65} & \underline{ 34.32 } /  40.06\;/ \textbf{16.15} & \textbf{47.14} & \textbf{80.67} / \textbf{63.71} & \textbf{71.38} / \textbf{87.98} & \textbf{87.00} & \textbf{71.02}& \textbf{50.50} & \textbf{53.20} & 29.07\% \\

DMAX \& Pseudo IMAX& \cellcolor[HTML]{D9D9D9}\underline{5.29} / \underline{6.69} & \cellcolor[HTML]{D9D9D9}\underline{3.22} / \underline{4.50} &\cellcolor[HTML]{D9D9D9}\textbf{35.43} / \textbf{41.30} / \underline{ 16.01} & \cellcolor[HTML]{D9D9D9}\underline{45.98} & \cellcolor[HTML]{D9D9D9}\underline{77.44} / \underline{54.72} & \cellcolor[HTML]{D9D9D9}\underline{67.29} / \underline{84.78} & \cellcolor[HTML]{D9D9D9}\underline{3.41} & \cellcolor[HTML]{D9D9D9}\underline{64.16} & \cellcolor[HTML]{D9D9D9}\underline{40.81} & \cellcolor[HTML]{D9D9D9}36.91 & \cellcolor[HTML]{D9D9D9}16.25\% \\ \bottomrule
\end{tabular}
\end{table*}

\subsubsection{Experiment Protocol} \hfill \\

To control for potential confounding factors arising from architectural designs, experiments are conducted under a unified MLLM architecture: EVA-CLIP ViT-G/14~\cite{fang2023eva} as the vision encoder; 2-layer multilayer perceptron (MLP) as the connector; LLaMA2-Chat (7B)~\cite{touvron2023llama} as the LLM.
During training, the visual backbone remains frozen with an input image resolution of 224*224 and the LLM is fine-tuned through LoRA~\cite{hu2022lora} with the rank of 8.
The training process lasts for 3 epochs with a linear warm-up phase spanning the first 10\% iterations.
The learning rate was configured with a peak value of 1e-6, decaying to 1e-7 through the cosine strategy.

Our experiments focus on two critical aspects:
(1) Characterizing the difference between using IMAX and DMAX in model training.
(2) Given the scarcity and high collection cost of IMAX data in practical scenarios, how can we develop an effective strategy to improve model performance using general DMAX data?

For the former, we randomly select 1,600 samples for each dataset.
Considering that the dimension of FIM is equal to the number of parameters $p$, directly calculating the eigenvalues is numerically intractable.
We adopt a simple rule where a single parameter is chosen per trainable layer, yielding the representative 450 parameters.
We set $N$ as 32 to calculate and observe statistical indicators that include $SE$ and $\rho$ at four checkpoints in the initial, early, mid, and late phases of the training process.

\begin{figure}[t]
\centerline{\includegraphics[scale=0.46]{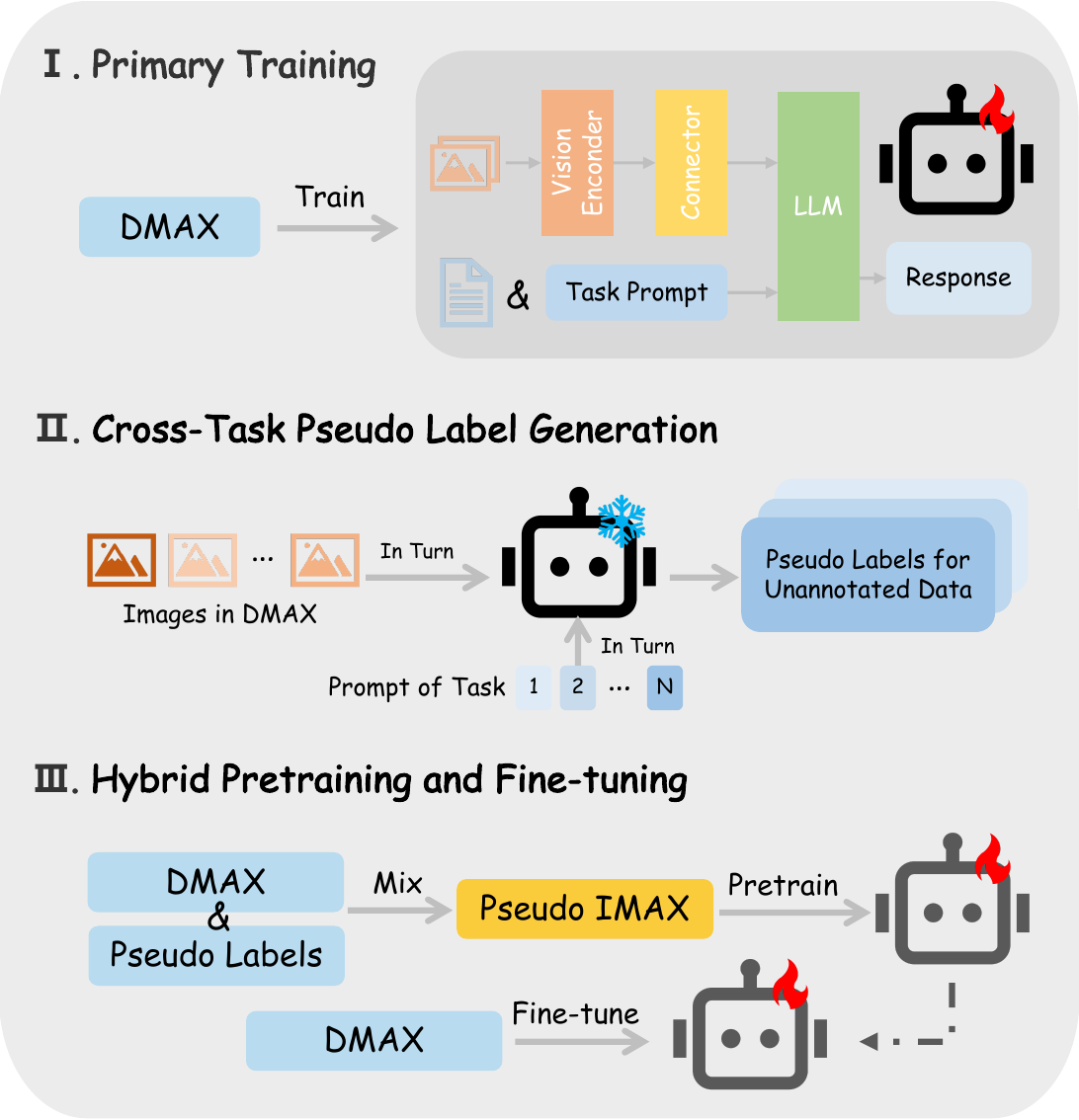}}
\caption{Illustration of our proposed training strategy for DMAX data.
Our pipeline comprises three stages: (1) primary training; (2) cross-task pseudo label generation; (3) hybrid pretraining and fine-tuning.}
\label{f4}
\end{figure}

To address the latter challenge, we draw inspiration from semi-supervised learning paradigms and propose a three-stage training strategy, as depicted in Figure~\ref{f4}.
To be specific, the process starts with MLLM training on the available DMAX data.
Then, the preliminary model engages in cross-task pseudo label generation, which leverages task-specific prompts to infer missing annotations for unlabeled data.
We can get pseudo IMAX data by integrating original DMAX with propagated pseudo labels.
The model subsequently undergoes hybrid pretraining om pseudo IMAX, followed by final fine-tuning on DMAX.

\subsubsection{Findings and Analyses}\hfill \\

\begin{figure}[h]
\centerline{\includegraphics[scale=0.52]{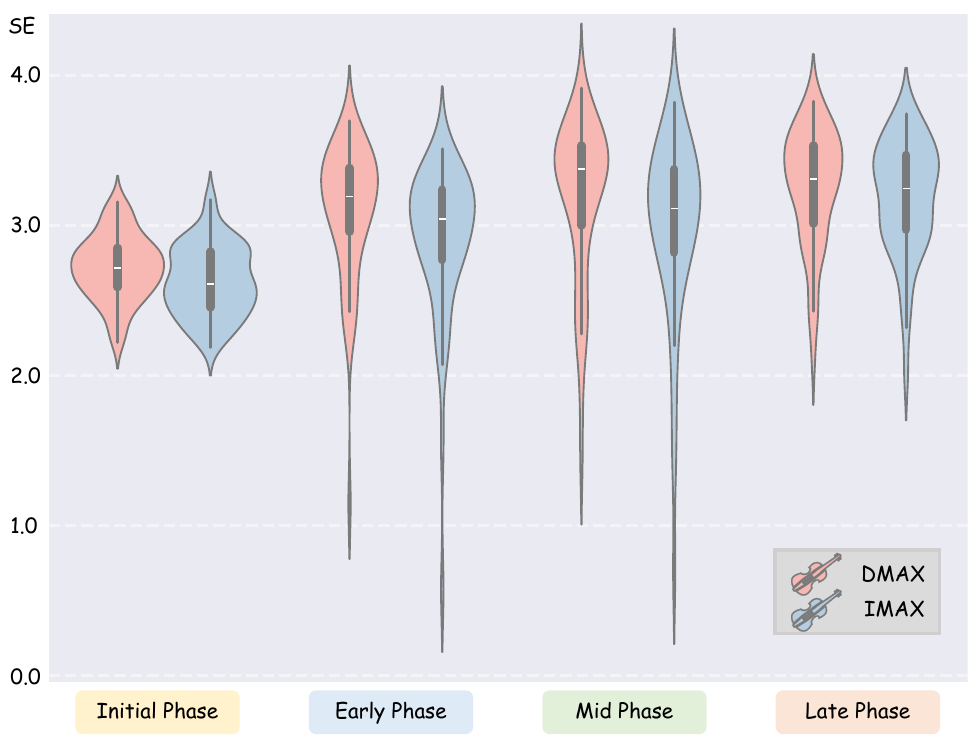}}
\caption{The violin plots and boxplots of the spectral entropy of the FIM eigenvalue at the initial, early, mid, and late phrases of training process.}
\label{f5}
\end{figure}

As shown in Figure~\ref{f5}, we implement a hybrid visualization approach that integrates violin plots and boxplots to examine the distributional characteristics of $SE$ from randomly selected train set samples at checkpoints in the initial, early, mid, and late phrases of the training process.
Our empirical observations reveal a statistically significant reduction in $SE$ during IMAX-based training compared to DMAX-based training, with average reductions of 3.38\%, 6.38\%, 5.80\% and 4.05\% at the four observation checkpoints, respectively.
In addition, we find consistent enhancements in the dominant eigenvalue ratio $\rho$ during IMAX-based training, with relative increases of 2.24\%, 4.62\%, 2.70\% and 2.19\%, respectively.

The first and second rows of Table~\ref{t2} demonstrate the evaluation metrics of the models trained with DMAX and IMAX, with the IMAX paradigm achieving substantial 29.07\% average performance gains.
Based on statistical evidence and evaluation metrics, we hypothesize that the observed pattern of reduced $SE$ and elevated $\rho$ during IMAX-based training indicates concentrated parameter optimization along critical directions, which may exhibit potential strong alignment with multi-task learning objectives, ultimately manifesting in significant improvements across evaluation metrics.

The third row of Table~\ref{t2} presents the results of our proposed DMAX-based training strategy.
Notably, when employing pseudo labels generated by the first row's model (exhibiting suboptimal test set performance), the pretraining with pseudo IMAX data confers average 16.25\% relative gains to DMAX-based fine-tuning, and even achieves competitive results compared to high-quality IMAX-based training on several individual metrics.
Our observation demonstrates that the core concept of image-centric multi-annotation data construction retains the capacity to exert the beneficial impact on multi-task learning even though data quality is relatively constrained.
Furthermore, under conditions of limited availability of high-quality IMAX data, our strategy shows enhanced compatibility with DMAX data while maintaining the balance between annotation quality and model performance, thereby significantly increasing practical utility in multi-task learning scenarios.

\section{Conclusion}
In this paper, we present IMAX, the first image-centric multi-annota-tion X-ray dataset designed to enhance the multi-task learning capabilities of medical generalist foundation models.
By addressing the critical limitation of decentralized image-task alignment in conventional multi-annotation data (e.g., DMAX), IMAX ensures comprehensive image understanding through dense annotations across seven different medical tasks.
Our experiments on seven open-source state-of-the-art medical MLLMs demonstrate that IMAX achieves 3.20\%–21.05\% average performance gains over DMAX, validating the effectiveness of image-centric data construction in fostering multi-task synergy.
Further statistics and analysis from the perspective of the Fisher information matrix reveals that IMAX training concentrates parameter updates along optimization directions with higher information content, ultimately leading to significant improvements in multi-task evaluation metrics.
Finally, to mitigate the practical challenges of acquiring high-quality IMAX data, we propose a three-stage DMAX-based training strategy built on the core principles of IMAX data construction, achieving the balance between annotation quality and model performance.
Our future directions include extending the concept of image-centric multi-annotation data to 3D medical modalities (e.g., CT and MRI) and exploring efficient multi-task learning of medical generalist foundation model in more complex mixed scenarios.
We will release related resources to promote community-wide advancements in generalist medical AI.

% \newpage
\bibliographystyle{ACM-Reference-Format}
\bibliography{sample-base}

\newpage
\appendix

\title{Supplementary Materials: \\
Enhancing Multi-task Learning Ability of Medical Generalist Foundation Model via Image-centric Multi-annotation Data}

\section{Data}

\subsection{Data Source}
\paragraph{Data for Calculation}
CXR-Cardiomegaly~\cite{duvieusart2022multimodal} provides image-derived cardiomegaly biomarker values for 96K chest X-rays in MIMIC-CXR-JPG.
The cardiothoracic ratio (CTR) is the most common measurement used by doctors to identify patients with cardiomegaly, which is defined as the ratio of the cardiac width (horizontal distance between extremes of the cardiac shadow) to the thoracic width (horizontal distance between the inner side of the ribs at the level of the hemidiaphragms).
Cardiopulmonary area ratio (CPAR) is defined as the ratio of the cardiac shadow area to the area of the lungs.
Both biomarkers are measured from posterior-anterior chest X-rays.
In clinical situations, CTR above 0.5 is considered to be pathological, while CPAR around 0.35-0.4 is appropriate.
For samples which are failed to extrapolate a CPAR or CTR value, there are 3 possible error markings: unable to locate heart, unable to locate lungs and unable to locate either.

\paragraph{Data for Report Generation}
MIMIC-CXR-JPG~\cite{johnson2019mimic} is a large dataset of chest radiographs with
free-text radiology reports.
A total of 377,110 JPG format images are available in the dataset from 227,827 image studies collected for 65,379 patients.
Each patient may have multiple studies and each study may contain one or more images associated with the same free-text report.

\paragraph{Data for Multi-class Classification}
MIMIC-CXR-PE-Severity~\cite{horng2021deep} provides pulmonary edema severity grades extracted from 6710, 485, and 141 X-rays from the MIMIC-CXR-JPG dataset through different means: 1) by regular expression from radiology reports, 2) by expert labeling from radiology reports, and 3) by consensus labeling from chest radiographs.
The extracted pulmonary edema severity labels are numerically coded as follows: 0, none; 1, vascular congestion; 2, interstitial edema; and 3, alveolar edema.

\paragraph{Data for Multi-label Classification}
The first iteration of CXR-LT provides a large multi-label, long-tailed CXR dataset.
Each image is labeled with one or multiple labels from a set of 26 disease findings by parsing radiology reports.
The newest version of CXR-LT~\cite{holste2024towards} extract labels for an additional 19 rare disease findings (for a total of 377,110 CXR images, each with 45 disease labels).
In our study, we select 10 disease labels including lung opacity, pleural effusion, atelectasis, pneumonia, edema, consolidation, pneumothorax, fracture, infiltration and nodule for the task of multi-label classification.

\paragraph{Data for Referring Expression Comprehension}
CheXmask~\cite{gaggion2024chexmask} aggregates 657,566 anatomical segmentation masks derived from images in five public databases: ChestX-ray8, Chexpert, MIMIC-CXR-JPG, Padchest and VinDr-CXR.
To construct consistent and high-quality data for the task of referring expression comprehension, we only use segmentation masks with a mean dice reverse classification accuracy (RCA) greater than 0.7.
The segmentation masks of the objects is formatted in run-length encoding (RLE).
We calculate the bounding boxes, i.e. the minimum bounding rectangle, for each mask based on the original image resolution, and store them in [$X_{min}$, $Y_{min}$, $X_{max}$, $Y_{max}$] format.
We normalize the coordinates to the range of 0 to 1000 in practical use.

\paragraph{Data for Referring Expression Generation}
The target of referring expression generation is exactly opposite to referring expression comprehension.
We still use CheXmask~\cite{gaggion2024chexmask} to construct the corresponding data with completely consistent rules.

\paragraph{Data for Visual Question Answering}
MIMIC-Ext-MIMIC-CXR-VQA~\cite{bae2023ehrxqa} is a complex, diverse, and large-scale dataset designed for the visual question answering task within the medical domain, focusing primarily on chest radiographs.
This dataset includes approximately 377K entries derived from the MIMIC-CXR-JPG, MIMIC-IV, and Chest ImaGenome datasets.
MIMIC-Ext-MIMIC-CXR-VQA defines a total of 48 templates, all of which are evaluated by a medical expert for clinical importance.
These templates fall into 3 semantic types (defining the response required: verify for yes/no questions; choose for selection from options; query for open-ended questions) and 7 content types (classifying the question's focus: presence, anatomy, attribute, abnormality, size, plane, and gender).

\subsection{Raw Data availability}
In the Table~\ref{sm1}, we list the links for each dataset.
All raw data in this study are publicly available, adhere to strict ethical standards, and undergo thorough anonymization with identifiable details removed.

\begin{table*}[]
\small
\caption{Some Typical Commands}
\label{sm1}
\begin{tabular}{lll}
\toprule
Task Type & Dataset & Link \\
\midrule
Calculation & CXR-Cardiomegaly & {https://physionet.org/content/cxr-cardiomegaly/1.0.0/} \\
Report Generation & MIMIC-CXR-JPG & {https://www.physionet.org/content/mimic-cxr-jpg/2.1.0/} \\
Multi-class Classification & MIMIC-CXR-PE-Severity & {https://www.physionet.org/content/mimic-cxr-pe-severity/1.0.1/} \\
Multi-label Classification & CXR-LT & {https://cxr-lt.github.io/CXR-LT-2024/} \\
Referring Expression Comprehession & CheXmask & {https://www.physionet.org/content/chexmask-cxr-segmentation-data/1.0.0/} \\
Referring Expression Generation & CheXmask & {https://www.physionet.org/content/chexmask-cxr-segmentation-data/1.0.0/} \\
Visual Question Answering & MIMIC-Ext-MIMIC-CXR-VQA & {https://physionet.org/content/mimic-ext-mimic-cxr-vqa/1.0.0/} \\
\bottomrule
\end{tabular}
\end{table*}

\section{Medical MLLMs for Comparison}

We select seven open-source state-of-the-art medical MLLMs for extensive comparative analysis.

\paragraph{LLaVA-Med}
LLaVA-Med~\cite{li2023llava} is a biomedical vision-language assistant built upon LLaVA.
In the first stage of biomedical concept feature alignment, the model is aligned with biomedical visual concepts by training on 600K image-caption pairs filtered from the PMC-15M dataset, using simple instructions that ask the model to describe the image.
During this stage, both the vision encoder and LLM are kept frozen while only the linear projection layer is updated.
The second stage is end-to-end instruction-tuning, the model is fine-tuned on 60K instruction-following data generated from PMC-15M using GPT-4, covering multi-turn conversations across five common biomedical imaging modalities.
In this stage, only the visual encoder is frozen, while the projection layer and the LLM are trainable.
Finally, LLaVA-Med conduct fine-tuning on three biomedical VQA datasets including VQA-RAD, SLAKE, and PathVQA.
The repository of LLaVA-Med v1.0 provides the model weights after end-to-end instruction-tuning and fine-tuning to downstream datasets.

We use the weights of the second stage for initialization and conduct fine-tuning on our IMAX and DMAX, respectively.

\paragraph{XrayGPT}
As one of the pioneering works in conversational medical vision-language models, XrayGPT~\cite{thawakar2024xraygpt} can analyze and answer open-ended questions about chest radiographs.
It adopts a frozen medical visual encoder to extract visual features from X-ray images and a frozen Vicuna fine-tuned on 100K real medical conversations and 20K radiology-specific dialogues to capture domain-specific knowledge.
A learnable linear transformation layer is employed to align visual and textual modalities.
The training process involves two stages: 
(1) In the first stage, the model is pretrained to understand the relationship between X-ray features and their corresponding textual reports using 213,514 image-text pairs from the MIMIC-CXR dataset.
(2) In the second stage, the model is further trained on 3,403 carefully selected image-text pairs from the OpenI dataset to generate more fluent and high-quality radiology specific responses.
The radiology-tuned Vicuna and the checkpoints of the second stage have been released in the open-source repository.

Since original XrayGPT only makes connector parameters open during training, which is insufficient to support the multi-modal input-output alignment required for multi-task learning, we use the weights of the second stage for initialization and additionally unfreeze the LLM parameters.

\paragraph{MiniGPT-Med}
MiniGPT-Med is an MLLM designed for heterogeneous radiological vision-language tasks.
It is fine-tuned from stage 3 of MiniGPT-v2 using 124,276 medical images from MIMIC, NLST, SLAKE, RSNA, and RadVQA, covering X-rays, CT scans, and MRIs.
These images cover many different medical conditions and support a wide range of tasks, such as visual question answering, image captioning, referring expression comprehension and generation, disease detection, and grounded image captioning. 
During training, the vision encoder remains frozen.
The linear projection layer is fine-tuned directly, while the LLaMA-2 language model is fine-tuned using LoRA.
The final model weights are open source.

Following the configuration of the fine-tuning stage, we use the final weights of MiniGPT-v2 for initialization.

\paragraph{Uni-Med}
Uni-Med~\cite{zhu2024uni} is a medical generalist foundation model designed with a unified interface and shared parameters to support six heterogeneous medical tasks: question answering, visual question answering, report generation, referring expression comprehension, referring expression generation, and image classification.
To mitigate the tug-of-war problem, Uni-Med introduces CMoE, a well-designed replacement module for the connector of MLLMs.
During training, the visual backbone is kept frozen, CMoE is trained, while the LLM is fine-tuned using LoRA.
The training data consists of text and image-text pairs, with a total of 141,645 samples selected and processed from MedQA, PubMedQA, Path-VQA, Slake-VQA, MIMIC-CXR, MPx-Single, MedMNIST v2, and SA-Med2D-20M.

Uni-Med follows a single-stage training strategy and provides open-source code but not the model weights.
Therefore, we use the released code, randomly initialize the weights, and fine-tune the model separately on IMAX and DMAX.

\paragraph{LLaVA-Tri}
LLaVA-Tri~\cite{xie2024medtrinity} is an MLLM trained on the large-scale dataset MedTrinity-25M.
MedTrinity-25M is constructed through an automated pipeline that leverages MLLMs to generate multigranular image-ROI-description triplets from unpaired medical images.
These triplets provide both global annotations, such as disease type, modality, and inter-region relations, and local annotations, including bounding boxes, segmentation masks, and region-specific descriptions.
The model is first pretrained on 600K image-text pairs from PMC-15M, then further trained on MedTrinity-25M for multi-granular alignment, and finally fine-tuned on the VQA-RAD, SLAKE, and PathVQA, respectively.
All fine-tuned versions of the model have been open-sourced.

Since the weights trained only on PMC-15M and MedTrinity-25M were not released, and only the weights further fine-tuned on downstream tasks are available, we initialize the model with the checkpoint fine-tuned on the smallest dataset, SLAKE, to approximate the state before downstream task tuning.

\paragraph{MedM-VL}
MedM-VL investigate various architectural designs following the encoder–connector–LLM of LLaVA and adopts a two-stage training strategy involving multi-modal pretraining and instruction tuning.
In the first stage, only the connector is trained to align image and text using image-caption-style data, where LLaVA’s original dataset is used for 2D tasks and CT-RATE reports are used for 3D tasks, respectively.
In the second stage, the entire model is trained to improve performance on various vision-language tasks.
MedM-VL-2D for 2D medical image analysis uses SigLIP as the image encoder, a two-layer MLP as the connector, and Qwen2.5-3B as the language model, while MedM-VL-CT-Chest for 3D CT-based applications uses M3D-CLIP as the 3D image encoder.
The 2D training data include Path-VQA and Slake-VQA for visual question answering, MIMIC-CXR and MPx-Single for report generation, MedMNIST v2 for image classification, and SA-Med2D-20M for referring expression, while the 3D training data come from CT-RATE.
Both MedM-VL-2D and MedM-VL-CT-Chest are available in the open-source repository.

Since Med-VL has not released the weights after the first stage of multi-modal pretraining, we use the second stage weights of MedM-VL-2D for initialization.

\paragraph{HealthGPT}
HealthGPT~\cite{lin2025healthgpt} is a medical MLLM that integrates medical visual comprehension and generation capabilities through a three-stage learning strategy and a novel H-LoRA technique.
% stage1
In the first stage, abstract- and concrete-grained visual adapters are trained separately for comprehension and generation.
For medical comprehension tasks, the abstract-grained adapters are trained to align visual and text embeddings, while the LLM and its H-LoRA modules stay frozen.
For generation tasks, the concrete-grained adapters and H-LoRA modules are trained instead, while the LLM remains unchanged.
% stage2
In the second stage, the shared word embedding layer and output head of the H-LoRA sub-modules are fine-tuned to ensure that the multiple H-LoRA plugins seamlessly interface with the LLM and form a unified base.
% stage3
In the third stage, H-LoRA modules and adapter modules are trained using additional task-specific data to enhance the model's adaptability to downstream tasks such as medical QA, report generation, super-resolution, and modality conversion.
% setting
The model uses CLIP-L/14 for model's dynamic hierarchical visual perception and phi-3-mini or phi-4 as the base model.
Training is based on the VL-Health dataset, which includes PubMedVision, LLaVA-Med, PathVQA, MIMIC-CXR-VQA, SLAKE, VQA-RAD, open world data from LLaVA-1.5, LLaVA-558k, IXI, and SynthRAD2023.
% open-source
HealthGPT has released the weights after the third stage, including the phi-3-mini version (HealthGPT-M3) for both comprehension and generation, as well as the phi-4 version (HealthGPT-L14) for comprehension.

We initialize the model using the HealthGPT-M3 and HealthGPT-L14 weights for comprehension and then perform fine-tuning, respectively.

%%
%% The next two lines define the bibliography style to be used, and
%% the bibliography file.

\end{document}